\documentclass[twocolumn,10pt]{jmlr} 

\usepackage{booktabs}
\usepackage{siunitx}
\theorembodyfont{\upshape}
\theoremheaderfont{\scshape}
\theorempostheader{:}
\theoremsep{\newline}

\title[Downstream Fairness Caveats with Synthetic Healthcare Data]{Downstream Fairness Caveats with Synthetic Healthcare Data}

\author{%
\Name{Karan Bhanot}\Email{bhanotkaran22@gmail.com}\\
\addr Rensselaer Polytechnic Institute, Troy, New York, USA
\AND
\Name{Ioana Baldini}\Email{ioana@us.ibm.com}\\
\Name{Dennis Wei}\Email{dwei@us.ibm.com}\\
\addr IBM Research, Yorktown Heights, New York, USA
\AND
\Name{Jiaming Zeng}\Email{jiaming@ibm.com}\\
\addr IBM Research, Cambridge, Massachusetts, USA
\AND
\Name{Kristin P. Bennett}\Email{bennek@rpi.edu}\\
\addr Rensselaer Polytechnic Institute, Troy, New York, USA
}

\begin{document}

\maketitle

\begin{abstract}
This paper evaluates synthetically generated healthcare data for biases and investigates the effect of fairness mitigation techniques on utility-fairness. Privacy laws limit access to health data such as Electronic Medical Records (EMRs) to preserve patient privacy. Albeit essential, these laws hinder research reproducibility. Synthetic data is a viable solution that can enable access to data similar to real healthcare data without privacy risks. Healthcare datasets may have biases in which certain protected groups might experience worse outcomes than others. With the real data having biases, the fairness of synthetically generated health data comes into question. In this paper, we evaluate the fairness of models generated on two healthcare datasets for gender and race biases. We generate synthetic versions of the dataset using a Generative Adversarial Network called HealthGAN, and compare the real and synthetic model's balanced accuracy and fairness scores. We find that synthetic data has different fairness properties compared to real data and fairness mitigation techniques perform differently, highlighting that synthetic data is not bias free.
\end{abstract}

\section{Introduction}
\label{sec:intro}
Accessing healthcare data can lead to real-world solutions such as using Electronic Medical Records (EMRs) for better patient care and improve clinical research \citep{cowie2017electronic}, analyse and model diseases \citep{10.1145/3292500.3330718} and more. Even though healthcare data is abundant, access to it is often restricted by privacy laws such as Health Insurance Portability and Accountability Act (HIPAA) in the United States \citep{HIPAA} and General Data Protection Regulation (GDPR) in the European Union \citep{GDPR}. Albeit essential, these laws hinder reproducibility and limit research. As a result, access to healthcare data records requires data de-identification (e.g. Stanford OMOP \citep{omop}), contracts and other methods to protect private information. Even so, there are risks of information loss, data breaches and leaks, with over 40 million individuals impacted due to data breaches just in 2021 as reported to United States Department of Health and Human Services (HHS) \citep{mckeon_2021}. These can lead to huge penalties with an average healthcare data breach cost of \$9.42 million per incident \citep{hipaa_journal_2021}.

Synthetic healthcare data addresses the problem of limited access by generating data which resembles the real data while preserving patient privacy and thus, can be disseminated without restriction. Synthetically generated data can be released with published research, enabling the community to reproduce existing research and conduct new research using these datasets. Synthetic data has been used across many applications, such as in the use of synthetic ``difference in differences'' in place of conventional estimators \citep{arkhangelsky2021synthetic}, generating Monte-Carlo simulations \citep{ATHEY2021}, etc. Within the healthcare domain, current research has evaluated synthetic data for resemblance, privacy, and utility \citep{yalethesis, 10.1007/978-3-030-61146-0_26}, showing it to be effective for categorical, binary and continuous datasets \citep{10.1007/978-3-030-61146-0_26} and time-series datasets \citep{dash2020medical}. This can be especially impactful during pandemics such as COVID-19 where widespread research can help but data access may be limited, for example, a study found that synthetic COVID-19 data had similar characteristics to the real data and could be used as a good proxy \citep{el2021evaluating}. However, healthcare datasets can have biases themselves. For example, certain groups defined by protected attributes may experience worse outcomes than others. In \citep{doi:10.1146/annurev.publhealth.012809.103613}, the study found that American Indians or Alaska Natives fared worst as a group in terms of relative health disparities when compared with other groups.

With real healthcare data being potentially biased, the fairness of synthetic healthcare data also comes into question. In healthcare, we want the synthetic data to be an accurate representation of the real data while preserving patient privacy, such that any analysis on the synthetic data is replicable on the real data. This implies that the synthetic data should capture the characteristics of the real data, having similar utility and fairness, even when the data may be biased. Further, application of any fairness mitigation should ideally resemble if it were to be applied on the real data. This shall make the use of synthetic data as a proxy of the real data robust.

Synthetic data is considered to have good ``utility'' when a ML model generated on it performs similar to if the same ML model was generated on the real data. While any classification metric can be used to measure utility, such as accuracy and f1\_score, we use balanced accuracy to measure the performance of the models for all possible class predictions. Simultaneously, while fairness can take many forms, we define ``fairness'' as measured using group fairness metrics including equal opportunity difference, average odds difference and equalized odds on the generated ML models. These fairness metrics measure the performance of the model using True Positive Rate (TPR) and False Positive Rate (FPR) difference between the sub groups of a protected attribute such as gender, ethnicity, age etc. 

In this paper, we measure the utility-fairness trade-off between real and privacy-preserving synthetic healthcare datasets by generating Machine Learning (ML) models. To provide a comprehensive analysis, we measure two biases: (a) ``gender bias'' as defined as the Female vs Male bias problem and (b) ``race bias'' as defined as the Black vs White bias problem. We evaluate the gender bias in the Cardiovascular dataset \citep{cardiovascular} and the race bias in the MIMIC-3 dataset \citep{johnson2016mimic} as measured by fairness metrics on Random Forest (RF) models. We also apply three fairness mitigation techniques to understand if unfairness can be mitigated similarly for the models on the real and synthetic datasets. This paper specifically aims to answer the following questions:
\begin{itemize}
    \item Is synthetic data similar to the real data as measured by utility and fairness? 
    \item Do fairness mitigation techniques work similarly on the real and synthetic data? 
    \item What is the utility-fairness trade-off between real and synthetic datasets?
\end{itemize}

\section{Related work}

While there are several ways to generate synthetic data, Generative Adversarial Networks (GANs) \citep{goodfellow2014generative} have been effective in the synthetic generation of many datasets, and are one of the most promising solutions. They have been used for image generation \citep{Karras_2019_CVPR}, image-to-image translation \citep{Huang_2018_ECCV, Zhu_2017_ICCV}, text generation \citep{10.1007/978-981-15-4409-5_71}, image generation based on text descriptions \citep{Zhang_2017_ICCV}, music generation \citep{briot2021artificial}, and financial time-series generation \citep{doi:10.1080/14697688.2020.1730426}. Advanced versions of GANs have been compared with other generative methods such as stacked Restricted Boltzmann Machines (RBMs), Variational Autoencoders and have demonstrated better performance \citep{medGAN}.

It has been shown with experiments on Boundary Equilibrium GAN (BEGAN), Deep Convolutional GAN (DCGAN) and DCGAN with Variational AutoEncoders (DCGAN-VAE) that attackers can perform membership inference attacks on GANs \citep{hayes}. This is worsened in the collaborative environment, where one study was able to use an inference attack to identify/generate samples of the training data \citep{DBLP:journals/corr/HitajAP17}. In the healthcare domain, privacy of the patients is of the utmost importance and hence, synthetic data generators must be privacy preserving. Such models include DPGAN \citep{DBLP:journals/corr/abs-1802-06739}, medGAN \citep{medGAN}, HealthGAN \citep{10.1007/978-3-030-61146-0_26, yale2020generation} etc. DPGAN provides theoretical guarantees on privacy preservation but the observed results show reduced utility. medGAN measured the privacy preservation but is only limited to count and binary datasets. HealthGAN is a model which has shown promising results on both public and private healthcare datasets catering to wide range of data types and thus, is our candidate model for fairness evaluation.

Synthetic data generators are not free from biases. In \cite{9666131}, the authors demonstrate the biases that exist in data generated by GANs by evaluating the distributions of generated images. An in-depth comparison of representativeness between real and synthetic data, some based on time-series metrics for synthetic data introduced earlier \citep{timeseries_esann}, has shown that synthetic data represents different sub-groups unfairly \citep{bhanot2021problem}. While this work studied and quantified representative unfairness, the effect of synthetic data on utility and associated fairness in downstream tasks was not addressed. \citep{10.1145/3442188.3445879} studied image datasets and found that differentially private GANs led to lower utility but showed inconclusive trends in group fairness metrics. Another study on census data observed variable biases in different synthetic models using several fairness metrics \citep{gupta2021transitioning} but its results were restricted to non-healthcare datasets. In contrast, our work focuses on tabular healthcare data, discusses real vs. synthetic fairness-utility trade-offs and also investigates how fairness mitigation techniques affect these trade-offs.

\section{Methodology}
\label{sec:methodology}

\subsection{Synthetic Data Generation}
Synthetic healthcare data involves the generation of data that imitates real data without revealing real patient details. Synthetic data generated by Generative Adversarial Networks (GANs) have shown promising results. HealthGAN has especially been shown to produce high utility and resemblance patient data while preserving patient information across a number of research studies \citep{yale2020generation, dash2020medical}. Motivated by the high-quality results, we decided to use the HealthGAN model for synthetic data generation using the official repository \citep{synthetic_data_repo}. 

We generated the synthetic datasets for two healthcare datasets. The first dataset is the Cardiovascular dataset \citep{cardiovascular} which includes information collected during medical examination of patients such as age, gender, blood pressure, blood pressure etc and has been studied before \citep{TORFI2022485}. The dataset includes information about whether the patient has a cardiovascular disease or not and is used as the label for ML models. We use gender as the protected attribute with values as Female or Male. The second dataset is an excerpt of the Multiparameter Intelligent Monitoring in Intensive Care (MIMIC)-3 dataset \citep{johnson2016mimic} derived based on a previous study to explore how 30-day mortality is affected by race and the associated synthetic datasets \citep{mundkur2017use, yalethesis}. Race is considered as the protected attribute in this data which can take five values: Asian, Black, Other, Unknown and White.

\begin{figure}[htbp]
\floatconts
  {fig:experiment_structure}
  {\caption{Process of data splitting, synthetic data generation, ML model training and evaluation.}}
  {\includegraphics[width=\linewidth]{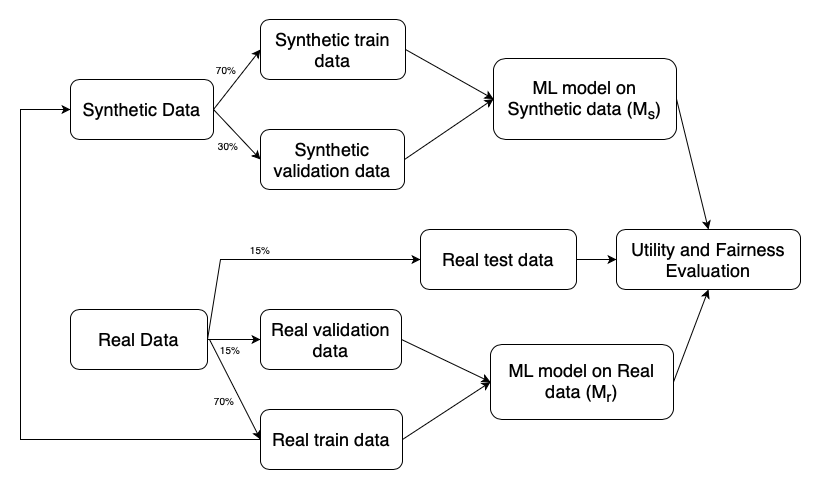}}
\end{figure}

We performed a train-test split with 70-30 ratio of the real data and used the training part of the real data to generate and evaluate the synthetic data. The schematic is shown in Figure~\ref{fig:experiment_structure}. Each data generation includes generation of 100,000 records. To accommodate variations across different train-test splits, we repeat the experiment for 10 runs. Each time a seed value from 1 to 10 (based on the experiment run) is selected to split the real data, and the resultant training data is used for synthetic data generation. The generator was tested for different epochs for best resemblance and privacy scores: 50K, 100K and 200K. We set epochs to 100K for Cardiovascular dataset and 200K for MIMIC-3 dataset, selected to achieve the best resemblance and privacy scores as defined by the nearest neighbors Adversarial Accuracy (nnAA) metric \citep{yale2020generation}.

\subsection{Machine Learning (ML) model training}

In the Cardiovascular dataset, there are outliers which are highly unlikely to occur. Thus, these are removed after data generation but before model training. Firstly, the age is in days so it is converted to years by diving by 365.25 followed by one-hot encoding of the gender column. Next, blood pressure values greater than 360 and less than 20 are also removed which led to better classification results. Random Forest (RF) models was trained on the datasets. The model was searched across a range of possible values using grid search for best balanced accuracy. The RF model achieved the best validation balanced accuracy with 100 trees and maximum features set to 10 with maximum depth set to auto.

In the MIMIC-3 dataset, the data is one-hot encoded for all columns as they are all categorical. We again train RF models on the datasets, with parameters identified using grid search to achieve best validation balanced accuracy. RF is trained with 20 trees, each with a maximum depth of 5 and maximum features automatically identified by the algorithm.

The models are trained using the scikit-learn library \citep{scikit-learn} in Python. The models trained on real and synthetic data are denoted $M_r$ and $M_s$ respectively. For identifying the best results in terms of balanced accuracy scores, we traverse a range of threshold values, 0.01 to 0.50 with 0.01 step, to determine the threshold that achieves the highest balanced accuracy (utility). We repeat the experiment 10 times, each time changing the split by updating the seed and present the results across all splits. The models $M_r$ and $M_s$ are evaluated on the real-test data with the complete experiment structure described in Figure~\ref{fig:experiment_structure}. We refer to these models designed for achieving the best balanced accuracy as the ``baseline'' models.

We also trained Logistic Regression (LR) models for the two datasets, identified using grid search for best validation accuracy scores. However, we found that due to the complex nature of the datasets, LR models did not always converge. Thus, the LR model was not able to learn the patterns in the datasets well, especially during the application of fairness mitigation techniques. Hence, the results based on the RF model provide the best understand on the results for the considered datasets and will be the focus of this paper.

\subsection{Fairness metrics}
For the trained models, we measure three ML fairness metrics: Equal Opportunity Difference, Average Odds Difference and Equalized Odds, to evaluate if the models perform equally well across sub-groups of protected attributes between real and synthetic data. We use the AI Fairness 360 (AIF360) library \citep{bellamy2018ai} by IBM to evaluate the fairness of the models. We briefly define each metric below with their mathematical definitions made available in \appendixref{apd:fairness_metrics}.

\textbf{Equal Opportunity Difference} measures the difference in the True Positive Rate (TPR) between the subgroups of a protected attribute \citep{bellamy2018ai}. The model is considered fair within a range -0.1 to 0.1 \citep{aif360demo}.

\textbf{Average Odds Difference} measures the average of the differences in the False Positive Rate (FPR) and the True Positive Rate (TPR) between the subgroups of the protected attribute \citep{bellamy2018ai}. The model is considered fair within a range -0.1 to 0.1 \citep{aif360demo}.

\textbf{Equalized Odds} measures the maximum of the absolute value of the differences in the False Positive Rate (FPR) and the True Positive Rate (TPR) between the subgroups of the protected attribute \citep{DBLP:journals/corr/HardtPS16} with a fair value at 0. Similar to the two metrics above, we define the fairness of a model based on equalized odds to be between 0 and 0.1

In healthcare datasets, it is possible that sub-groups of protected attributes experience different rates of diagnosis and mortality. For example, Turner syndrome is a sex chromosome condition found only in females \citep{sutton2005turner}. As a result, mortality due to it can be non-zero only for females but not for males.  As a result, fairness metrics such as demographic parity that aim for equal rates across sub-groups do not always work for evaluating the ML models. Thus, in this paper, we selected the metrics to measure how well do the models perform based on their predictions for various subgroups, measured using both TPR and FPR. TPR is important as correctly identifying the individuals who have high probability of mortality or being diagnosed with a disease, can enable healthcare practitioners to direct appropriate resources to these patients. Similarly, FPR ensures that we do not provide excessive care to patients who don't need it, at the cost of other patients when resources are limited. The three metrics provide outlook at TPR and FPR from different perspectives.

\subsection{Fairness mitigation techniques}
We apply fairness mitigation techniques to see whether and by how much unfairness in synthetic healthcare data can be mitigated as compared to the real data. The following mitigation techniques are used to combat unfairness.

\textbf{EO Thresholding} tunes the classification threshold to achieve better equalized odds, similar to traversing the thresholds for achieving the highest balanced accuracy. Thus, as the first mitigation technique, we traverse the threshold values to achieve the lowest Equalized Odds (EO) with the condition that the balanced accuracy is at least 58\%. This ensures that, even while striving for better fairness, we do not compromise by considerably dropping utility.

\textbf{Reweighing} is a pre-processing technique that modifies the weights of various samples such that the data is more fair before generating a ML model \citep{kamiran2012data}. We apply reweighing to the training dataset before the LR model is trained, using the implementation in AIF360 \citep{bellamy2018ai}. During the model training, we include the weights identified using reweighing with the \textit{instance\_weights} parameter provided for models in scikit-learn.

\textbf{Reduction} is an in-processing technique that reduces a fair classification problem into a sequence of cost-sensitive classification problems \citep{DBLP:journals/corr/abs-1803-02453}. To better allow the prediction of probabilities and not just binary-valued predictions, we use the Grid Search Reduction method available in Microsoft's Fairlearn library \citep{bird2020fairlearn}.

\textbf{HPS} is the acronym we use (based on the authors' names) to refer to the post-processing technique introduced at the same time as the equalized odds fairness criteria \citep{DBLP:journals/corr/HardtPS16}. HPS proposes an optimization that minimizes equalized odds difference while also minimizing classification loss. We used the open-source implementation available in AIF360 \citep{bellamy2018ai}.

We trained ML models and applied these techniques for both real and synthetic data. As both reweighing and reduction allow for the choice of threshold, we use the EO thresholding criteria above to further improve fairness scores. We then compare results from all models for fairness metrics and balanced accuracy scores.

\section{Results}
\label{sec:results}

\subsection{Real vs. Synthetic Prevalence Rates}
\label{sec:results:prevalence}

Prevalence rates measures the proportions of individuals who belong to that sub-group in comparison to the whole population. We calculate the prevalence rates for the protected attributes in the two datasets and average the rates across 10 runs. The percentage change from real to synthetic data is also calculated and all results are shown in Table~\ref{tab:prevalence_rates_cardio} and Table~\ref{tab:prevalence_rates_mimic3} for Cardiovascular and MIMIC-3 datasets respectively. We applied the paired t-test \citep{Ross2017} using the scipy package \citep{2020SciPy-NMeth} between the rates in the real and synthetic data. The test has a null hypothesis that the mean difference within the pairs (each split gives rise to a pair) is zero while the alternative hypothesis states otherwise. 

\begin{table}[hbtp]
\setlength{\tabcolsep}{3pt}
\floatconts
  {tab:prevalence_rates_cardio}
  {\caption{Mean Prevalence Rates For Sub-Groups of Protected Attributes in Cardiovascular Data (*= statistically significant using paired t-test)}}
  {\begin{tabular}{ccccc}
    \toprule
    Sub-group & Real & Synthetic & Change & T-test \\
    & rate & rate & & p-value \\
    \midrule
    Female & 65.04 & 65.54 & 0.78\% & 2.12e-02* \\  
    Male & 34.96 & 34.46 & -1.45\% & 2.12e-02* \\
    \bottomrule
  \end{tabular}}
\end{table}

\begin{table}[hbtp]
\setlength{\tabcolsep}{3pt}
\floatconts
  {tab:prevalence_rates_mimic3}
  {\caption{Mean Prevalence Rates For Sub-groups Of Protected Attributes in MIMIC=3 Data (*= statistically significant using paired t-test)}}
  {\begin{tabular}{ccccc}
    \toprule
    Sub-group & Real & Synthetic & Change & T-test \\
    & rate & rate & & p-value \\
    \midrule
    Female & 48.48 & 48.64 & 0.32\% & 2.00e-01\\  
    Male & 51.52 & 51.36 & -0.3\% &2.00e-01 \\
    \midrule
    Asian & 2.76 & 3.30 & 19.53\% & 2.43e-02*\\
    Black & 9.89 & 6.53 & -34.02\% & 6.04e-12*\\
    Other & 6.86 & 7.16 & 4.37\% & 6.30e-02\\   
    Unknown & 9.32 & 7.53 & -19.17\% & 5.40e-05*\\
    White & 71.17 & 75.48 & 6.06\% & 7.01e-09*\\ 
    \bottomrule
  \end{tabular}}
\end{table}

\begin{figure*}[htbp]
\floatconts
  {fig:real_vs_synth_eq_opp_diff}
  {\caption{Scatter plots comparing the real and synthetic fairness and utility scores where each line joins the real data and its corresponding synthetic data. (a) Comparison of equal opportunity difference and balanced accuracy between real and synthetic Cardiovascular dataset while measuring gender bias (b) Comparison of equal opportunity difference and balanced accuracy between real and synthetic MIMIC-3 dataset while measuring race bias}}
  {\includegraphics[width=\linewidth]{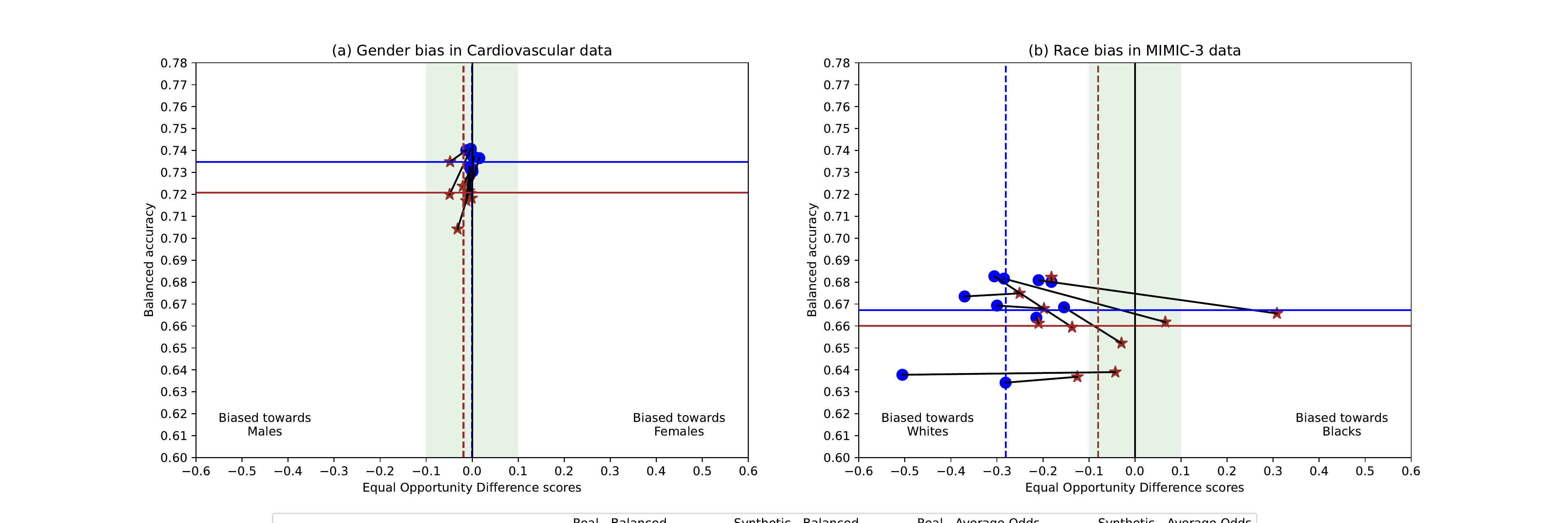}}
\end{figure*}

The results underscore that certain sub-groups might be under-represented and others might be over-represented in the synthetic data, indicating a shift in the distribution of the data. We find that the change in gender prevalence rates between real and synthetic data is significant (p-value $<$ 0.05) for the Cardiovascular data but not for MIMIC-3 (p-value $>$ 0.05). However, all race subgroups in MIMIC-3, except Other, had p-value $<$ 0.05. This means the difference is significant and we can reject the null hypothesis. We find that Black and Unknown ethnic sub-groups are under-represented in the synthetic data, Blacks being the most affected by over 30\%. This is in contrast to the White, Asian and Other sub-groups which have all seen an over-representation. Thus, for a comprehensive analysis, we compare the utility and fairness scores for gender bias in the Cardiovascular data and race bias in the MIMIC-3 data. We select the Black vs White problem for race bias as they create a highly skewed data with more than 70\% population being White while Blacks experiencing the worst change in prevalence in synthetic data.

\subsection{Real vs Synthetic Fairness-Utility}

Figure~\ref{fig:real_vs_synth_eq_opp_diff} compares the equal opportunity difference and balanced accuracy scores between the real and synthetic data for the RF models. The subplot on the left highlights the gender bias in the Cardiovascular dataset while the subplot on the right shows the race bias in the MIMIC-3 data. Here, the balanced accuracy and fairness are measured only on the sub-groups selected for the comparison rather than the whole dataset.

In the Cardiovascular data, the real data as measured for gender bias is quite fair with the mean equal opportunity difference value very close to 0. However, as we move from the real data to the synthetic data, we note that the unfairness increases as indicated by the dashed vertical maroon line being farther away from 0. The average balanced accuracy also decreases with some synthetic data models experiencing worse drops than others. While visually the differences are not extreme, statistically the difference in equal opportunity difference values ($p\_val = 0.0073$) and balanced accuracy ($p\_val = 0.0006$) between real and synthetic data are significant. Overall, there is a decrease in both utility and fairness of the data.

\begin{figure*}[htbp]
\floatconts
  {fig:fairness_mitigation_avg_odds_diff}
  {\caption{Box plots showing the range of Average Odds Difference values for models generated on various datasets and application of fairness mitigation techniques, for (a) Gender bias measured on Cardiovascular dataset and (b) Race biased measured on MIMIC-3 dataset. The varios models are: best accuracy model on real data, EO thresholding on real data, reweighing on real data, reduction on real data, HPS on real data, best accuracy model on synthetic data, EO thresholding on synthetic data, reweighing on synthetic data, reduction on synthetic data and HPS on synthetic data}}
  {\includegraphics[width=\linewidth]{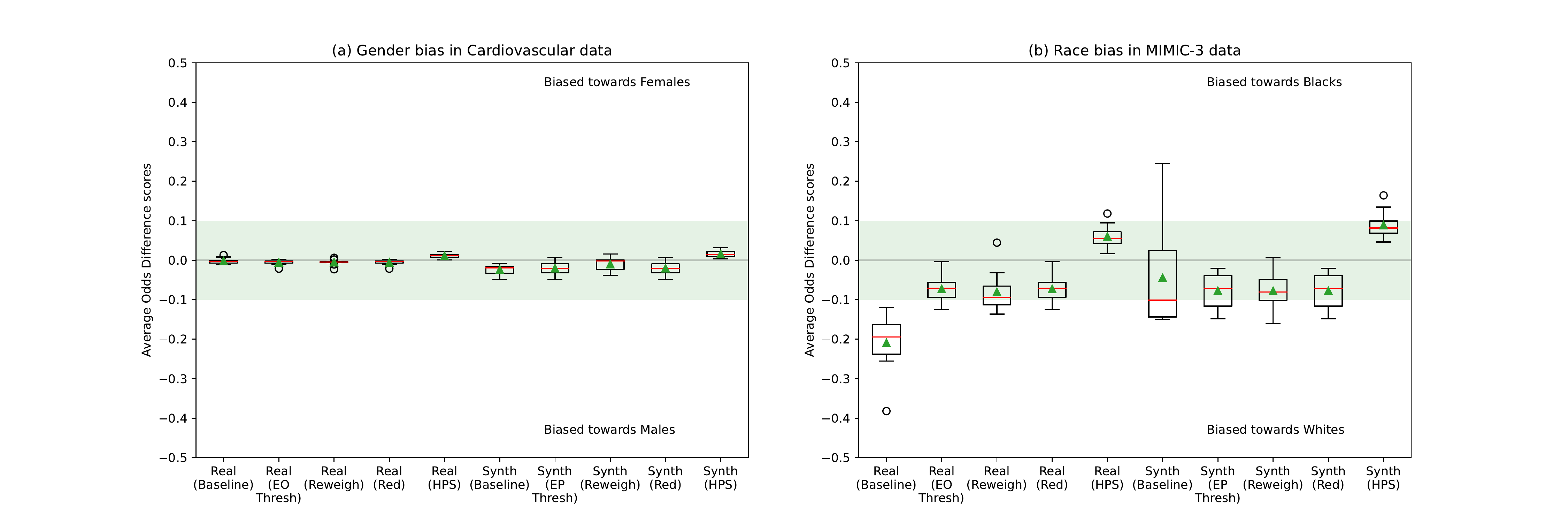}}
\end{figure*}

In the MIMIC-3 data, we observe that the real data is quite biased towards the White population as indicated by the dashed blue vertical line. In synthetic data models, we observe that there is an improvement in fairness in almost all cases, moving from region of unfairness to fairness. However, in some cases, the fairness points moved from a region of being biased towards Whites to a region biased towards Blacks. Although synthetic data makes the results fairer as indicated by the maroon dashed vertical line, it actually introduced bias non-existent in the real data. Thus, even when the fairness is improved, the synthetic data is not useful. These differences are also statistically significant ($p\_val = 0.0067$). Even though the average balanced accuracy decreased, the differences are not statistically significant ($p\_val = 0.0563$). Thus, we conclude that the race bias present in the real data is not replicated in the synthetic data, rather synthetic data introduced new biases.

Similar plots for average odds difference and equalized odds along with a table of paired t-test results are included in \appendixref{apd:real_synthetic_fairness}. These scores between real and synthetic data are also statistically significant. The results underscore that synthetic data has fairness different from the real data. Thus, if this synthetic data is used as is for conducting research, we might see results which are different from the real data. Even though the results are fairer than the real data when evaluating race bias in MIMIC-3, introducing biases in the opposite direction is not the ideal goal and thus, needs to be identified and rectified before this synthetic data is released for widespread use.

\subsection{Fairness Mitigation}
We apply four fairness mitigation techniques: EO thresholding, reweighing, reduction and HPS, on both the real and synthetic data to understand how they would improve the fairness of the models. These are then compared by generating box plots for the 10 runs of the experiment. The results for equal opportunity difference for the RF models are described in Figure~\ref{fig:fairness_mitigation_avg_odds_diff}. The left subplot shows the range of equal opportunity difference scores between females and males while measuring gender bias in Cardiovascular data. The right subplot shows the range of equal opportunity difference scores between blacks and whites while measuring race bias in MIMIC-3 data. While the fairness mitigation techniques target for equalized odds, we discuss the average odds difference results as the existence of positive and negatives values shows if and how any improvements are compromising fairness towards a specific subgroup.

For gender bias in Cardiovascular data, we find that fairness mitigation techniques lead to small changes in the real data. Reduction ($p\_val = 0.0461$) and HPS ($p\_val = 0.0071$)lead to statistically significant results than the real baseline. In synthetic data, we find that the variance is much more than the real data. EO thresholding, reweighing and reduction all lead to increased variance with only reweighing leading to better mean and median fairness scores with statistically significant difference ($p\_val = 0.0336$). HPS on synthetic data changes the bias from males to females while also being statistically significant ($p\_val = 0.0002$) compared with its synthetic baseline. Thus, we find that the mitigation techniques do not work equally well on both the real and synthetic data, sometimes introducing non-existent biases.

For race bias in MIMIC-3 data, we clearly see the difference in the real and synthetic data baselines as discussed in the previous subsection. The real data is biased towards Whites while the synthetic data, even though fairer, is biased towards Whites in some cases and Blacks in others, introducing new biases. Fortunately, application of fairness mitigation techniques lead to improvements on both the real and synthetic data. EO thresholding, reweighing and reduction improve fairness while HPS reverses the direction of fairness on the real data. These differences are also statistically significant, indicating a definite change in fairness. Similar results are also observed in the synthetic data as well, however, the results are not statistically significant except for HPS ($p\_val = 0.0224$) compared to the synthetic baseline.

We observe similar results for equal opportunity difference and equalized odds with their corresponding results included in \appendixref{apd:fairness_mitigation_results}. We find that fairness mitigation techniques do lead to improvements in fairness scores for both the real and synthetic datasets. However, the extent of improvement varies and is sometimes not significant. Thus, we observe that different mitigation techniques work for real and synthetic dataset in the two cases considered. Furthermore, similar technique when applied on real and synthetic data does not result in similar improvements.

\subsection{Utility-Fairness Trade-off}
To understand the effect of fairness mitigation techniques on utility as measured by balanced accuracy, we created scatter plots between the fairness metrics and balanced accuracy. Figure \ref{fig:mitigation_fairness_vs_utility_eq_odds} shows the equalized odds and balanced accuracy plots. The plot is split into 4 subplots, the first two corresponding to gender bias in Cardiovascular data while the last two corresponding to the race bias in MIMIC-3 data for the RF models.

\begin{figure*}[htbp]
\floatconts
  {fig:mitigation_fairness_vs_utility_eq_odds}
  {\caption{Scatter plots comparing the Equalized Odds and balanced accuracy scores for real data with synthetic data and various fairness mitigation techniques: (a) Gender bias in Cardiovascular real data, (b) Gender bias in Cardiovascular synthetic data, (c) Race bias in MIMIC-3 real data and (d) Race bias in MIMIC-3 synthetic data. The mean values of Equalized Odds and balanced accuracy are shown using vertical dashed and horizontal solid lines respectively for each corresponding technique and dataset.}}
  {\includegraphics[width=\linewidth]{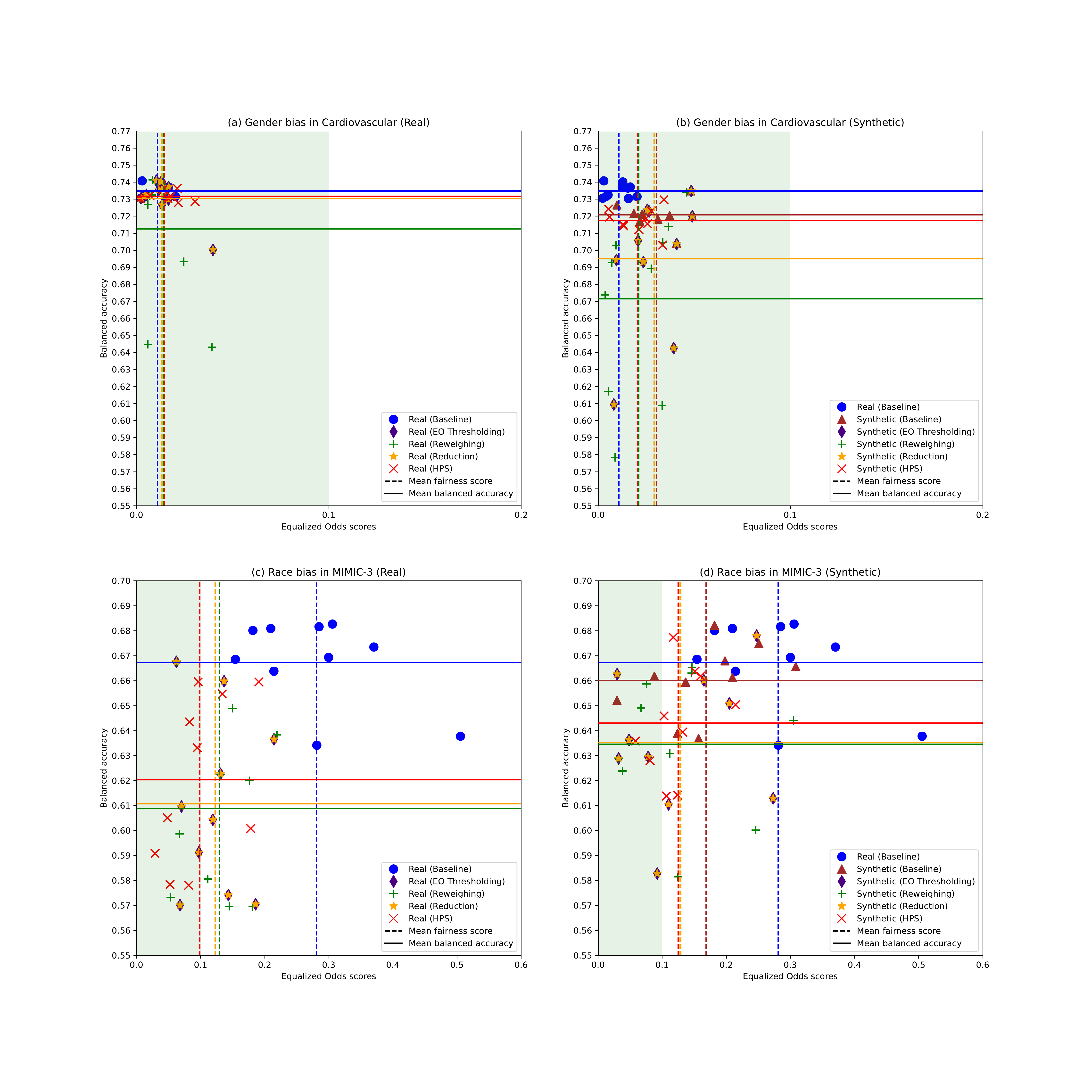}}
\end{figure*}

For gender bias in Cardiovascular dataset, we observe that application of all fairness mitigation techniques on the real data led to a decrease in equalized odds and balanced accuracy. The furthest drop was observed in reweighing, indicating that it not only led to increased unfairness but also worsened the performance of the model in terms of utility. However, these differences are not statistically significant as measured by paired t-test except for balanced accuracy difference observed in HPS ($p\_val = 0.0028$). In synthetic data, even though all mitigation techniques led to lower balanced accuracy scores, they did so for improving the fairness of the model. Reweighing led to improvements in fairness but at the cost of the most drop in balanced accuracy. This is not ideal as the difference in equalized odds are not significant while the reduced balanced accuracy is statistically significant ($p\_val = 0.0156$). On the other hand, HPS improved fairness with statistical significance ($p\_val = 0.0110$) in comparison to the synthetic baseline while also not compromising on utility excessively. This difference in balanced accuracy of HPS in comparison to its synthetic baseline is also found to be statistically significant ($p\_val = 0.0015$).

For race bias in MIMIC-3 data, we observe that the fairness mitigation techniques perform significantly well across both the real and synthetic data. In the real data, HPS led to the most improvement in fairness while reducing the balanced accuracy the least. However, as seen through the results in the previous subsection, it introduces undesirable biases towards the black subgroup. On the other hand, EO thresholding, reweighing and reduction bring considerable improvements to fairness, however, at the cost of balanced accuracy. All these differences as measured for equalized odds and balanced accuracy scores are also statistically significant, indicating an actual improvement in fairness and simultaneous reduction in balanced accuracy. In the synthetic data, we find results similar to the real data. All fairness mitigation techniques led to improvements in fairness compared to both the real and synthetic baselines, however, at the cost of balanced accuracy. However, the relative improvements were minor in the synthetic data than the real data. The equalized odds improvements across all mitigation techniques are not statistically significant but the drop in balanced accuracies are.

Similar plots with average odds difference and equal opportunity difference are included in \appendixref{apd:fairness_mitigation_results}. From the results, we make two observations. In gender bias evaluation, application of fairness mitigation techniques did not lead to improvements in fairness for the real data but did so for synthetic data. In race bias evaluation, while all techniques led to improvements in both the real and synthetic data, the improvements in synthetic data were less (mean values are horizontally compressed). Thus, we observe that application of fairness mitigation techniques do not always guarantee an improvement in fairness. In cases when we do observe improvements, it is at the cost of reduced utility. Thus, we conclude that the utility-fairness trade-off is quite evident in these datasets, often influenced by the choice of protected attribute and real vs. synthetic data.
 
\section{Discussion}
\label{sec:discussion}
    
\subsection{Synthetic data has variable fairness}
To be useful in the real world, synthetic data should effectively capture the same degree of fairness or unfairness as in real data. 

On analysing the prevalence rates for the various sub-groups in the two datasets, we found that synthetic data was over-representing certain groups and under-representing others. On measuring the utility and fairness for specific sub-groups of protected attributes, we found variable results between real and synthetic data. While evaluating gender bias in Cardiovascular dataset, we found that synthetic data had poorer fairness and utility. When evaluating race bias in MIMIC-3 dataset, even though the fairness improved in the synthetic data, the results were not ideal. In some models generated on synthetically generated data, there were high biases towards the female population which do not exist in the real data. Further, the mean balanced accuracy was lower too.

We conclude that the utility of the synthetic data is almost always lower than the real data. However, this does not lead to any guaranteed improvements in fairness. The results highlight that synthetic data generated for these datasets have less predictable movements in fairness and may create new biases. Thus, usage of these synthetic data without proper fairness evaluation can lead to solutions with unknown and undesired biases, rendering the results less than ideal for the real-world.

\subsection{Real vs Synthetic fairness mitigation is not same}
We conclude that application of fairness mitigation techniques often lead to improvements on both the real and synthetic data when the original data is unfair. However, the extent of observed improvements are often different.

As noted in gender bias analysis in Figure~\ref{fig:fairness_mitigation_avg_odds_diff}, reduction and HPS led to statistically significant differences with the baseline for real data but reweighing and HPS had statistically significant differences with the baseline in the synthetic data. In contrast, all fairness improvements were significant for the real data but none for synthetic data in race bias analysis. 

The fairness mitigation results highlight an important observation: Not only may synthetic data have undesired biases which are not present in the real data, but fairness mitigation techniques may be less effective as well. This implies that the same technique results in lower improvements on synthetic data, making it less ideal for current real-world use.

\subsection{Utility reduction does not mean fairness improvement}
From the results in Figure~\ref{fig:mitigation_fairness_vs_utility_eq_odds}, we clearly see that application of any fairness mitigation technique leads to a reduction in balanced accuracy. However, such a reduction does not always lead to better fairness scores. Furthermore, a higher reduction in utility also does not guarantee a more fairer model.

The mitigation results suggest a trade-off between utility-fairness which is often exacerbated by synthetic data. Achieving higher fairness by using fairness mitigation techniques on these dataset often comes at the cost of reduced utility. These observations and discussions are specific to the patient cohorts considered but highlight that synthetic datasets are not bias free. Thus, based on the healthcare application and the required trade-offs, fairness metrics and mitigation techniques should be evaluated before data release.

\section{Conclusion}
\label{sec:conclusion}

Previously, synthetic healthcare data has been evaluated for utility, resemblance and privacy \citep{yalethesis}. However, as healthcare data is often biased, it is important to evaluate the synthetic data for fairness as well. In this paper, we generate synthetic data for two healthcare datasets and measure balanced accuracy and ML fairness metrics. To understand how unfairness can be mitigated, we also experiment with fairness mitigation techniques.

We find that synthetic data suffers from poorer utility in comparison to the real data while also having variable movements in fairness, sometimes leading to undesired biases. Application of fairness mitigation techniques generally lead to improvements in fairness but their effect on real and synthetic data is different. Some improvements in synthetic data from its baseline are also not statistically significant to be useful for fairness improvement. Further, these mitigation techniques often lead to reduced utility without guaranteeing an improved fairness environment.

We conclude that the synthetic data must be evaluated for fairness such that its real-world applicability can be identified. This suggests that data generation need to be improved. Data generators could incorporate a controlled way to handle fairness, such that the generated data is generated with known biases. Models similar to the representative and fair synthetic data generator \citep{tiwald2021representative} can be extended to the healthcare domain to generate fair data with high utility and resemblance, and controlled fairness and privacy. This future direction of work can lead to the development of data generators which can capture the real data fairness well and use it as a means for controlled fair healthcare data generation.

\bibliography{jmlr-sample}

\appendix

\section{Fairness Metrics}
\label{apd:fairness_metrics}

Below are the mathematical definitions of the three Machine Learning (ML) fairness metrics used in the paper: Equal Opportunity Difference (Equation~\ref{eq:equal_opp_diff}), Average Odds Difference (Equation~\ref{eq:avg_odds_diff}), and Equalized Odds (Equation~\ref{eq:equzlied_odds}). In the three definitions, True Positive Rate (TPR) is the ratio of correct predictions in the total predictions of the positive class and False Positive Rate (FPR) is the ratio of incorrect predictions of the positive class. In our study, a positive prediction (1) is when the model predicts that the patient died or has a cardiovascular disease. We define the two subgroups of a protected attribute as $sub1$ and $sub2$.

\begin{align}
    \label{eq:equal_opp_diff}
    equal\_opp\_diff = (TPR_{sub1} - \nonumber\\
                     TPR_{sub2})
\end{align}

\begin{align}
    \label{eq:avg_odds_diff}
    avg\_odds\_diff = \frac{1}{2}\Big(
                    (FPR_{sub1} - \nonumber\\
                    FPR_{sub2}) + \nonumber\\
                    (TPR_{sub1} - \nonumber\\
                    TPR_{sub2}) \Big)
\end{align}

\begin{align} \label{eq:equzlied_odds}
equalized\_odds = max \Big(
                |FPR_{sub1} - \nonumber\\
                FPR_{sub2}|, \nonumber\\
                |TPR_{sub1} - \nonumber\\
                TPR_{sub2}| \Big)
\end{align}

\section{Real vs. Synthetic Fairness}
\label{apd:real_synthetic_fairness}

We generate scatter plots comparing the balanced accuracy with fairness metrics for real and synthetic data across subgroups of gender in Cardiovascular dataset and race in MIMIC-3 dataset. Figure~\ref{fig:avg_odds_diff_real_vs_synth} and ~\ref{fig:eq_odds_real_vs_synthetic} show the change in average odds difference and equalized odds for the 10 runs of the experiment as we move from real data to the corresponding synthetic data.

\begin{figure*}[htbp]
\floatconts
  {fig:avg_odds_diff_real_vs_synth}
  {\caption{Scatter plots comparing the real and synthetic fairness and utility scores where each line joins the real data and its corresponding synthetic data. (a) Comparison of average odds difference and balanced accuracy between real and synthetic Cardiovascular dataset while measuring gender bias and (b) comparison of average odds difference and balanced accuracy between real and synthetic MIMIC-3 dataset while measuring race bias}}
  {\includegraphics[width=\linewidth]{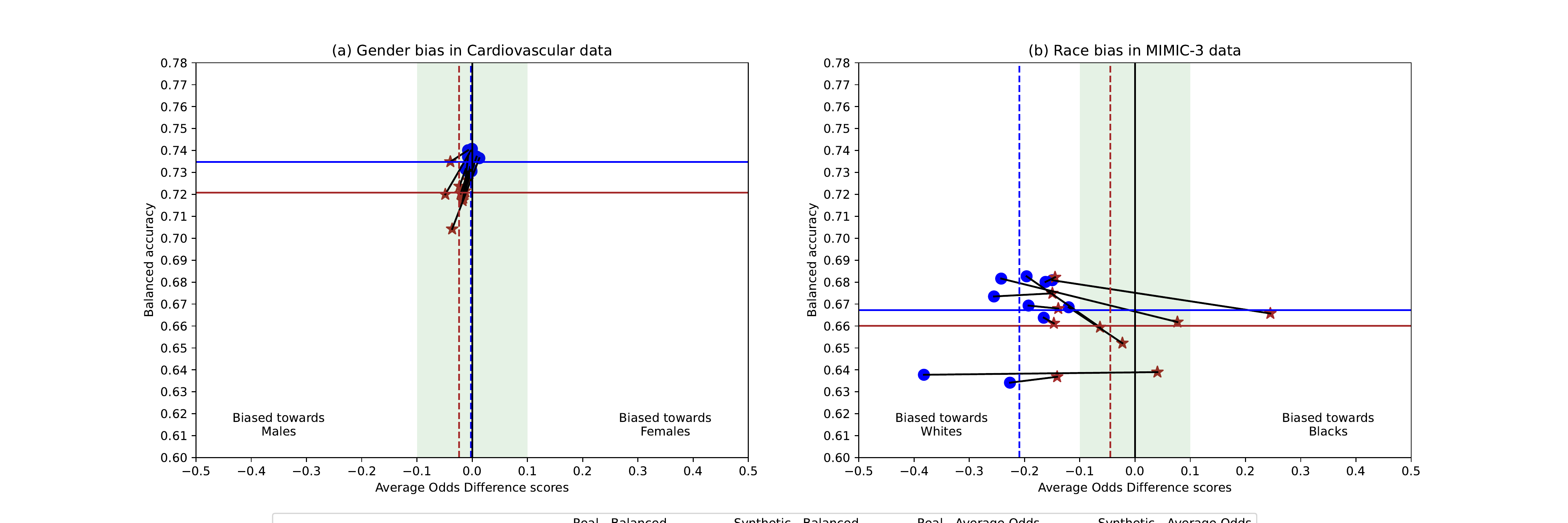}}
\end{figure*}

From Figure~\ref{fig:avg_odds_diff_real_vs_synth}, we observe that the results are similar to equal opportunity difference plots. The mean fairness and balanced accuracy for synthetic data are lower than the real data for gender bias. However, there are undesired biases in the direction of the black subgroup in the synthetic MIMIC-3 data being evaluated for race bias.

\begin{figure*}[htbp]
\floatconts
  {fig:eq_odds_real_vs_synthetic}
  {\caption{Scatter plots comparing the real and synthetic fairness and utility scores where each line joins the real data and its corresponding synthetic data. (a) Comparison of equalized odds and balanced accuracy between real and synthetic Cardiovascular dataset while measuring gender bias and (b) comparison of equalized odds and balanced accuracy between real and synthetic MIMIC-3 dataset while measuring race bias}}
  {\includegraphics[width=\linewidth]{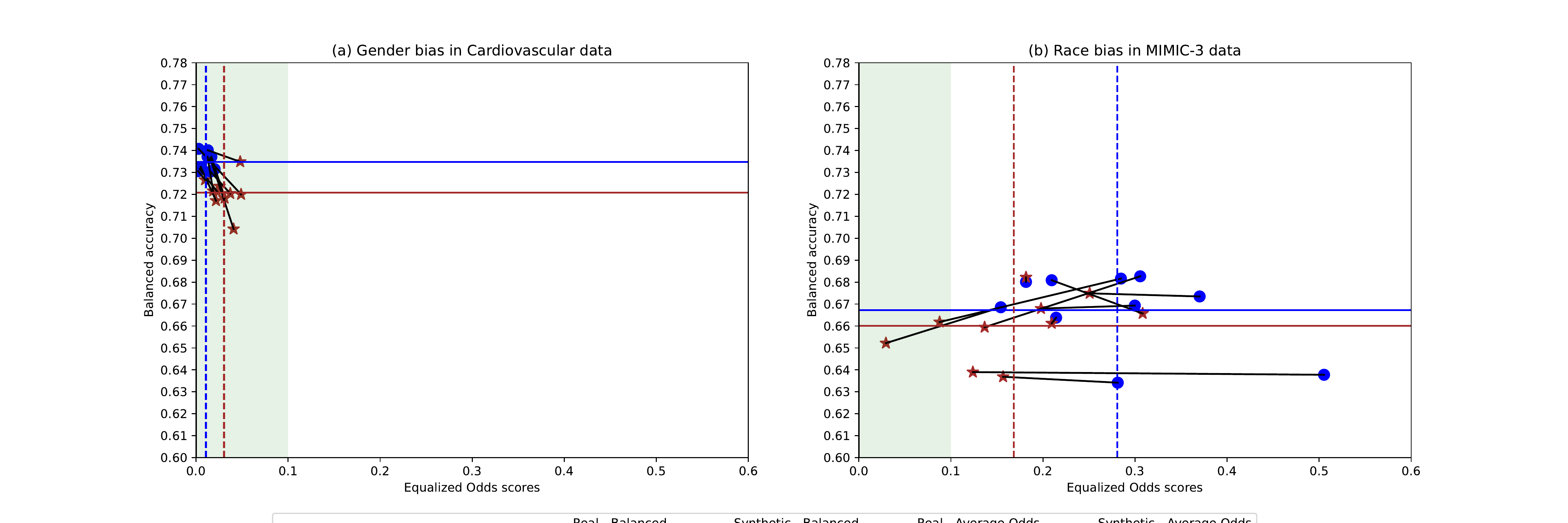}}
\end{figure*}

From Figure~\ref{fig:eq_odds_real_vs_synthetic}, we observe a similar trend in equalized odds. Fairness is worse for gender bias in Cardiovascular synthetic datasets while it is better for race bias in MIMIC-3 synthetic dataset compared to the real datasets.

\section{Fairness Mitigation Results}
\label{apd:fairness_mitigation_results}

The four mitigation techniques are applied on the real and synthetic datasets resulting in variable changes to utility and fairness. The fairness scores for equal opportunity difference are shown in Figure~\ref{fig:fairness_mitigation_eq_opp_diff} and the scores for equalized odds are shown in Figure~\ref{fig:fairness_mitigation_eq_odds}. The comparison of fairness scores with balanced accuracy values are shown in Figure~\ref{fig:mitigation_fairness_vs_utility_eq_opp_diff} and Figure~\ref{fig:mitigation_fairness_vs_utility_avg_odds_diff}. The Table~\ref{tab:fairness_mitigation_p_values} shows the p-values for various fairness and utility comparisons with respect to their corresponding baselines.

\begin{figure*}[htbp]
\floatconts
  {fig:fairness_mitigation_eq_opp_diff}
  {\caption{Box plots showing the range of equal opportunity difference values for models generated on various datasets and application of fairness mitigation techniques, for (a) Gender bias measured on Cardiovascular dataset and (b) Race biased measured on MIMIC-3 dataset. The various models are: best accuracy model on real data, EO thresholding on real data, reweighing on real data, reduction on real data, HPS on real data, best accuracy model on synthetic data, EO thresholding on synthetic data, reweighing on synthetic data, reduction on synthetic data and HPS on synthetic data}}
  {\includegraphics[width=\linewidth]{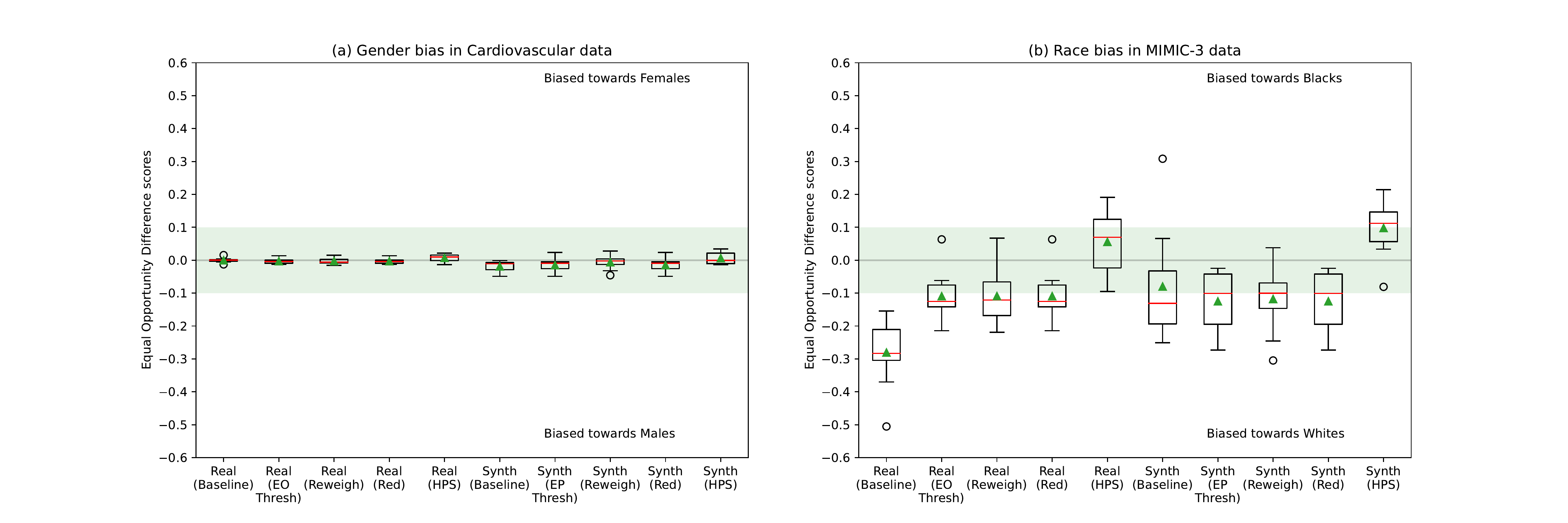}}
\end{figure*}

\begin{figure*}[htbp]
\floatconts
  {fig:fairness_mitigation_eq_odds}
  {\caption{Box plots showing the range of equalized odds values for models generated on various datasets and application of fairness mitigation techniques, for (a) Gender bias measured on Cardiovascular dataset and (b) Race biased measured on MIMIC-3 dataset. The various models are: best accuracy model on real data, EO thresholding on real data, reweighing on real data, reduction on real data, HPS on real data, best accuracy model on synthetic data, EO thresholding on synthetic data, reweighing on synthetic data, reduction on synthetic data and HPS on synthetic data}}
  {\includegraphics[width=\linewidth]{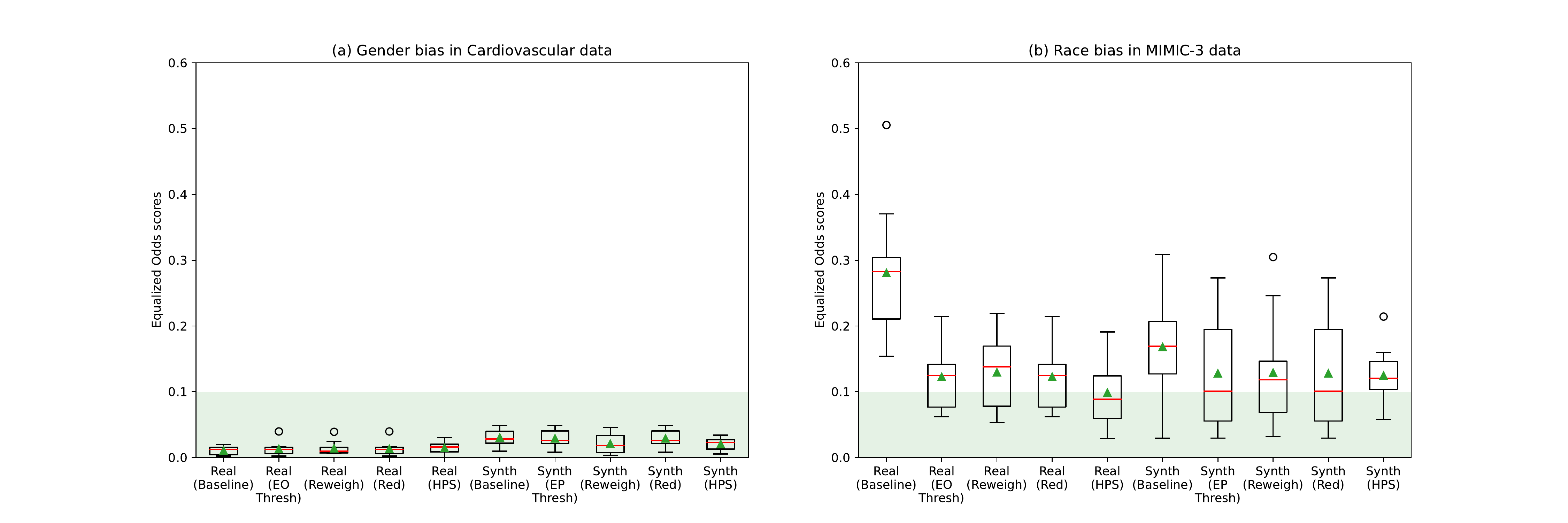}}
\end{figure*}

\begin{figure*}[htbp]
\floatconts
  {fig:mitigation_fairness_vs_utility_eq_opp_diff}
  {\caption{Scatter plots comparing the equal opportunity difference and balanced accuracy scores for real data with synthetic data and various fairness mitigation techniques: (a) Gender bias in Cardiovascular real data, (b) Gender bias in Cardiovascular synthetic data, (c) Race bias in MIMIC-3 real data and (d) Race bias in MIMIC-3 synthetic data. The mean values of equal opportunity difference and balanced accuracy are shown using vertical dashed and horizontal solid lines respectively for each corresponding technique and dataset.}}
  {\includegraphics[width=\linewidth]{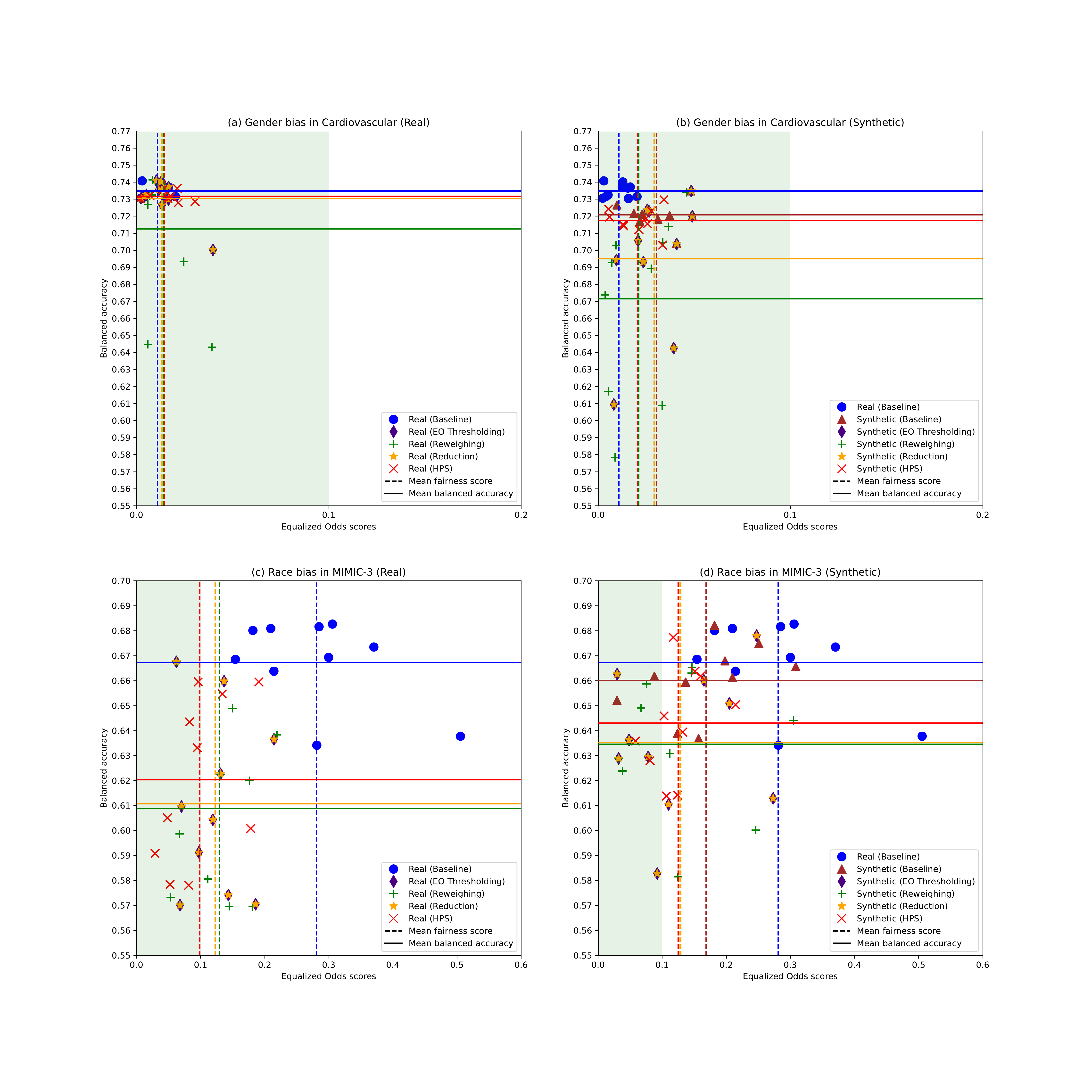}}
\end{figure*}

\begin{figure*}[htbp]
\floatconts
  {fig:mitigation_fairness_vs_utility_avg_odds_diff}
  {\caption{Scatter plots comparing the average odds difference and balanced accuracy scores for real data with synthetic data and various fairness mitigation techniques: (a) Gender bias in Cardiovascular real data, (b) Gender bias in Cardiovascular synthetic data, (c) Race bias in MIMIC-3 real data and (d) Race bias in MIMIC-3 synthetic data. The mean values of average odds difference and balanced accuracy are shown using vertical dashed and horizontal solid lines respectively for each corresponding technique and dataset.}}
  {\includegraphics[width=\linewidth]{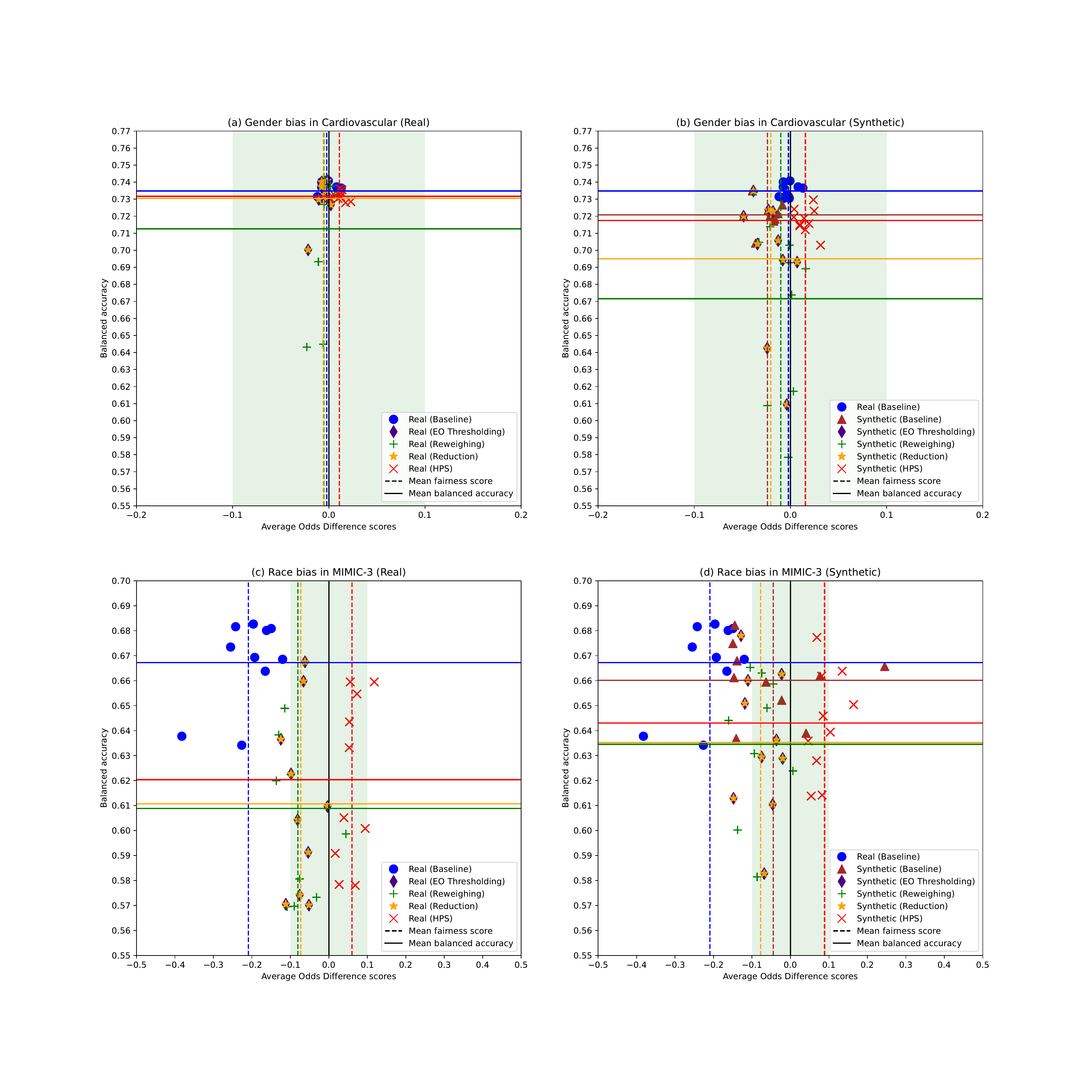}}
\end{figure*}

\begin{table*}[hbtp]
\floatconts
  {tab:fairness_mitigation_p_values}
  {\caption{Paired T-test P-values For Mitigation Techniques In Comparison To The Respective Baseline (*= statistically significant using paired t-test)}}
  {\begin{tabular}{cccccc}
    \toprule
    Bias & Fairness Mitigation & Bal Acc & Avg Odds & Equal Opp & Eq Odds \\
    \midrule
    Gender & EO Thresholding (Real) & 0.2034 & 0.0955 & 0.0461* & 0.2571 \\
    (Cardiovascular) & Reweighing (Real) & 0.0934 & 0.4935 & 0.2867 & 0.2054 \\
     & Reduction (Real) & 0.2034 & 0.0955 & 0.0461* & 0.2571 \\
     & HPS (Real) & 0.0028* & 0.187 & 0.0071* & 0.0584 \\
     & EO Thresholding (Synthetic) & 0.0702 & 0.1561 & 0.3418 & 0.7863 \\
     & Reweighing (Synthetic) & 0.0156* & 0.0192* & 0.0336* & 0.1516 \\
     & Reduction (Synthetic) & 0.0702 & 0.1561 & 0.3418 & 0.7863 \\
     & HPS (Synthetic) & 0.0015* & 0.0485* & 0.0002* & 0.011* \\
    \midrule
    Race & EO Thresholding (Real) & 0.0019* & 0.0001* & 0.0000* & 0.0001* \\
    (MIMIC-3) & Reweighing (Real) & 0.0022* & 0.0000* & 0.0001* & 0.0002* \\
     & Reduction (Real) & 0.0019* & 0.0001* & 0.0000* & 0.0001* \\
     & HPS (Real) & 0.0041* & 0.0001* & 0.0000* & 0.0004* \\
     & EO Thresholding (Synthetic) & 0.0076* & 0.4248 & 0.4485 & 0.3415 \\
     & Reweighing (Synthetic) & 0.0038* & 0.544 & 0.4668 & 0.3223 \\
     & Reduction (Synthetic) & 0.0076* & 0.4248 & 0.4485 & 0.3415 \\
     & HPS (Synthetic) & 0.0022* & 0.0431* & 0.0224* & 0.1229 \\
    \bottomrule
  \end{tabular}}
\end{table*}

\end{document}